\documentclass{sig-alternate}
\usepackage{times}
\usepackage{epsf}
\usepackage{amsmath}
\usepackage{amssymb}
\usepackage{graphicx}
\usepackage{times}
\usepackage{bm}
\usepackage{ascmac}

\begin{document}
%

\title{
Product Review Summarization based on \\ 
Facet Identification and Sentence Clustering
}

\numberofauthors{3} 
%
\author{
%
%
\alignauthor
Duy Khang Ly 
       \affaddr{National University of Singapore}\\
       \affaddr{Computing 1, \\13 Computing Drive}\\
       \affaddr{Singapore 117417}\\
       \email{ldkhang@gmail.com}
\alignauthor
Kazunari Sugiyama
       \affaddr{National University of Singapore}\\
       \affaddr{Computing 1, \\13 Computing Drive}\\
       \affaddr{Singapore 117417}\\
       \email{sugiyama@comp.nus.\\edu.sg}
\alignauthor
Ziheng Lin \\
       \affaddr{National University of Singapore}\\
       \affaddr{Computing 1, \\13 Computing Drive}\\
       \affaddr{Singapore 117417}\\
       \email{linzihen@comp.nus.\\edu.sg}
\and  
\alignauthor
Min-Yen Kan \\
       \affaddr{National University of Singapore}\\
       \affaddr{Computing 1, \\13 Computing Drive}\\
       \affaddr{Singapore 117417}\\
       \email{kanmy@comp.nus.\\edu.sg}
}

\maketitle

\begin{abstract}
 Product review nowadays has become an important source of information,
 not only for customers to find opinions about products easily
 and share their reviews with peers, but also for product manufacturers
 to get feedback on their products. 
 As the number of product reviews grows, it becomes difficult
 for users to search and utilize these resources in an efficient way. In this work,
 we build a product review summarization system that can automatically
 process a large collection of reviews and aggregate them to generate
 a concise summary. More importantly, the drawback of existing
 product summarization systems is that they cannot provide
 the underlying reasons to justify users' opinions. In our method,
 we solve this problem by applying clustering, prior to selecting representative candidates for summarization. 
\end{abstract}




\section{Introduction} \label{sec:Introduction}
 Product reviews are an important source of information.
 Not only do customers use them to find opinions about products,
 but it also allows them to vent their frustrations and
 share successes with their peers.  
 It also allows product manufacturers to receive feedback on their product lines. 
 Unfortunately, the number of reviews is
 overwhelming, making it difficult to search and utilize the resource.
 A user may not manage to read all relevant  reviews for a product 
  before needing to  make a decision on whether to  purchase it or not.
 The huge number of reviews also makes it difficult for product manufacturers
 to keep track of customer opinions of their products -- {\it e.g.},
 how do the public find about the recently released models,
 and what features do they expect to improve in the next models. 
 To address these issues, we build a product review summarization system
 that can automatically process a large collection of reviews and aggregate 
 information into a readable summary. Our system aims at 
 achieving the following two important goals:
 (1) to employ an efficient way to automatically identify topics
 and subtopics in the reviews (\textit{product facet identification}),
 and (2) to automatically summarize the correspondent opinions
 and present a coherent summary to users (\textit{summarization}).

 In (1) product facet identification,
 our approach first identifies frequent product dimensions
 being discussed in a review set. We show that the integration of
 a new heuristic using sentences' syntactic roles into one of
 the current state-of-the-art systems achieves better performance
 in precision. In (2) summarization, we implement a clustering
 algorithm that identifies a group of sentences sharing
 the same subtopic, before analyzing their sentiment and producing
 the desired output summary. Unlike previous approaches,
 the final summary is able to capture
 opinions from different dimensions of the product. More importantly,
 it allows a potential customer to quickly see how the existing 
 customers feel about the product, yet equip him/her with sufficiently
 detailed information. 

 This report is an extended version of \cite{Khang11}, which elaborates more on the approach used, and expands on the evaluation and analysis of our prior results.
 In Section \ref{sec:Related Work},
 we review related work on sentiment analysis and summarization.
 In Section \ref{sec:Proposed Method},
 we propose our product review summarization system.
 In Section \ref{sec:Experiments},
 we present the experimental results for evaluating our proposed approaches.
 Finally, we conclude the paper with a summary and directions for
 future work in Section \ref{sec:Conclusion}. 

\section{Related Work} \label{sec:Related Work}
 We divide the related work on the task of summarizing product reviews
 into two sub-fields: discovering the users' opinions
 expressed in the reviews (\textit{sentiment analysis}),
 and aggregating and arranging them in an appropriate output
 (\textit{summarization}).

\newpage
\subsection{Sentiment Analysis} \label{subsec:SA}
 Sentiment analysis refers to the computational treatment of subjectivity
 (whether there exists sentiment), the sentiment polarity
 (positive, negative, neutral or a scale of sentiment intensity),
 and the opinion content information (opinion holder, topic of opinion,
 {\it etc.}), that underlies a text span. The granularity of
 the text span starts at the level of individual words, then phrases, sentences,
 and finally the entire document.  These levels of granularity also offer a natural way of
 characterizing the techniques developed in sentiment analysis. However,
 we do not discuss work at the document level,
 as the target of our work is not to examine the overall sentiment of
 the review, but the detailed (and thus finer grained) opinions within the review.

 At the word level, Hatzivassiloglou and McKeown \cite{Hatz97}
 predicted the binary semantic orientation of adjectives.
 They utilized textual conjunctions (\textit{e.g.}, ``and,'' ``but'') 
 in a large training corpus between the target adjective and
 a seed list of adjectives with manual annotated polarity,
 achieving an accuracy of 82\% in average.
 Turney \textit{et al.}~\cite{Turney02} obtained comparable results
 with extended target words including not only adjectives,
 but also nouns, verbs and adverbs.
 Moreover, their system did not require a corpus as training data.
 Instead, they approximated the point-wise mutual information \cite{Church90}
 between the target word with the positive word ``excellent''
 and with the negative word ``poor,'' respectively, by counting
 the number of results returned by Web searches matching queries
 that join each pair of words by a NEAR operator. Since the scores
 correspond to the similarity between the target adjective with each
 positive/negative extreme, the polarity of that adjective can be
 determined by taking the label that results in the prominent score.
 More recently, Hu and Liu~\cite{Hu04} utilized WordNet~\cite{Miller95} --
 a large lexical database of English with synonym and antonym pointers
 -- to grow a initial seed list of known orientation
 adjectives into a larger list that covers all the remaining adjectives
 in WordNet. Their system achieved higher results (accuracy of 84\% 
 in average) than the two aforementioned systems,
 due to WordNet's stronger sense of organization compared with
 use of large text or Web corpora, as was used in the former two systems.

 The initial success of sentiment analysis at the word level provides
 the necessary building blocks for studying larger units of texts
 as shown in \cite{Wiebe99} and \cite{Bruce99}.
 Both works established a positive and statistically significant
 correlation with the presence of adjectives on determining
 the subjectivity of sentences, as well as documents. Furthermore,
 in determining the sentiment orientation of a sentence,
 Yu and Hatzivassiloglou~\cite{Yu03}, and Kim and Hovy~\cite{Kim04}
 aggregated the polarity of each individual adjective
 or sentimental word that appeared in the sentence itself.
 Following these works, Wiebe and Riloff~\cite{WiebeRiloff05},
 Wilson \textit{et al.}~\cite{Wilson05}, and Kim and Hovy~\cite{Kim06}   
 introduced additional sentence-surface features
 (\textit{e.g.}, counts of positive/negative adjectives
 in a target sentence, or in a window of previous
 and next sentences; binary feature on whether the sentence
 contains a pronoun, {\it etc.}) in a supervised manner,
 and then achieved fairly good results (up to an accuracy of 70\%)
 in the same task.

 Nevertheless, in the domain of product reviews,
 finding the orientation of the sentence is generally not enough.
 In fact, it is necessary to identify the semantics of
 the opinion in the sentence, as the opinion holder may
 describe a particular facet of the subject in the review that
 users may be interested in. Typical examples of facets that belong to
 a camera product would be: battery life, lens, flash system,
 price, and so on. In the case of a music player,
 the facets are: sound system, battery life, weight/size,
 storage capability, and so on. Hu and Liu~\cite{Hu04} addressed
 this problem by first applying data mining techniques to
 extract facets of the product, then classifying the orientation
 for each of the sentences where the facets appear in
 as positive or negative using WordNet. 
 Their system achieved promising accuracy of
 72\% in identifying product facets,
 and that of 84\% in predicting facet orientation. Subsequently,
 Popescu and Etzioni~\cite{Popescu05} introduced the use of
 relaxation labeling technique~\cite{Hummel83}
 in their OPINE system to determine facet orientation, and achieved
 an accuracy of 78\%.  They deem neighboring facets that appear in the
 same sentence as the target facet based on surface linguistic
 connective cues, such as conjunctions and disjunctions.  More
 recently, Ding \textit{et al.}~\cite{Ding08} proposed a
 state-of-the-art system that further incorporated a set of complex
 carefully-built grammar rules between adjacent sentence constructions
 as well as neighboring facets, together with a collection of
 comprehensive polarity-annotated lists of idioms, nouns, verbs,
 adjectives and adverbs, to solve the same problem. The system
 achieved an accuracy of 92\%, closely matching the upper bound of
 the performance of human perception.

 While the work on sentiment analysis discussed above make much of
 discovering the users' opinions in the reviews, few managed to
 aggregate these opinions together. In recent work, Sauper {\it et
   al.}~\cite{Sauper11} proposed an integrated approach that jointly
 learns product facets and user sentiments for product reviews 
 using Bayesian topic models. Another approach to this problem is to
 view the aggregation task as a summarization task, which we review next.

\subsection{Summarization} \label{subsec:RW-Summarization}
 In the early stages of the opinion summarization, Turney \textit{et
   al.}~\cite{Turney02} produced a thumbs-up/thumbs-down indication
 for movie reviews as the output of its orientation classification
 component.  The movie itself was treated as a single entity of
 interest.  Refining this to cater to the detailed characteristics of
 products, Hu and Liu~\cite{Hu04}, and Popescu and
 Etzioni~\cite{Popescu05} focused on product facets -- distinctive
 features of the product that users often make comments upon -- and
 generated facet-driven summary, supported with sentence-level
 statistics, \textit{i.e.}, the number of positive/negative sentences
 that the facet belongs to.  Subsequently, Liu \textit{et
   al.}~\cite{Liu05} extended the single facet-driven summary into a
 comparative-based summary between many products, where the
 orientation of all shared facets are plotted together with their
 number of supporting sentences for visualization. However, while
 users may prefer these systems for an at-a-glance presentation of
 products, they only provide only shallow information. In such
 systems, while users can learn that how many people prefer or dislike
 a facet, it does not explicitly help users organize the (shared)
 underlying reasons for their opinions.

 Multi-document summarization techniques are more relevant, since the
 task does not address a single review but a set of reviews.
 The main characteristic of multi-document
 summarization is both leveraging and cleaning up the inherent redundancy of the
 input, where similar information often appears across different
 sources.  Dejong~\cite{Dejong82} as well as Radev and
 McKeown~\cite{Radev98} applied information extraction techniques to
 gather information from different sources, and generated summaries by
 filling those extracted information into some predefined sentence
 templates.  However, their frameworks require significant background
 knowledge in order to create the detailed templates at a suitable
 level, and this fact results in domain-dependent system.  Barzilay
 \textit{et al.}~\cite{Barzilay99} proposed a novel approach that does
 not depend on domain-specific knowledge.  In their system, each
 sentence is first transformed into a predicate-argument structure
 called a DSYNT tree~\cite{Kittredge83} with the nodes being the
 sentence constituents. Under this representation, grammar
 dependencies between sentence constituents (subject-verb relation,
 adjective-noun relation, {\it etc.}) are captured and essentially
 abstracted from their ordering in the sentence.  Therefore, with the
 assistance of a set of paraphrasing rules that are capable of
 recognizing identical or similar predicates, they were able to derive
 rules to combine similar DSYNT trees of sentences from different
 sources together. The resulting tree is fed to a final sentence
 generation component to formulate a new sentence.  Carbonell and
 Goldstein's maximum marginal relevance (MMR)~\cite{Carbonell98} is
 another widely used technique in multi-document summarization;
 for example, Ye \textit{et al.}~\cite{Ye05} 
 leveraged MMR to solve their summarization task on general
 news to obtain reasonable results. In details, MMR is
 an iterative algorithm, which selects a sentence
 from the collection per round to insert into the final summary
 based on: (1) the selected sentence
 covering the most new information mentioned by the remaining
 unselected sentences, and (2) the selected sentence also has minimum similarity
 with all previously selected sentences in the summary.
 The algorithm terminates either when a fixed number of
 sentences is selected, or when the content overlapping
 between any candidate sentence and the summary
 at that iteration exceeds a predefined threshold.

\subsection{Shortcomings of Related Work} \label{subsec:Shortcomings}
 As described in Section~\ref{subsec:SA}, there exists two systems
 \cite{Hu04} and \cite{Popescu05} that addressed the problem of
 product facet identification. However, these systems only analyze users' opinions
 in the review and do not summarize these opinions. 
 Furthermore, it is not clear how they constructed queries
 that combine a set of cue words associated with the product
 class (\textit{e.g.}, ``of camera,'' ``camera has,''
 ``camera comes with,'' and so on) and the candidate facet together.
 Our own early experiments with different
 query combinations also do not show consistent results with their systems. 
 In recent work, Titov and McDonald~\cite{Titov08} proposed a joint
 statistical model to find the set of relevant facets for a rated
 entity and extracted all textual mentions that are associated with each
 other. But they focus on finding the set of facets and do not
 tackle summarization.

 We can see that the works of \cite{Turney02,Hu04,Popescu05,Liu05}
 focus on sentiment analysis rather than summary generation, but
 do not address the problem of extracting the underlying reasons for an
 opinion.
 To solve this
 problem, in this paper, we apply summarization techniques to produce
 user-friendly product review output.  Multi-document
 summarization~\cite{Dejong82,Radev98,Barzilay99}, techniques that
 previously experimented on news, have yet to be adapted for the
 domain of product reviews.  Product reviews differ from news articles
 in that they may not be grammatically well-formed and crucially, involve
 sentiment analysis.  In ~\cite{Ye05}, the applied MMR variant
 requires a metric to compute the content similarity between any two
 sentences, but when it comes to our domain of product reviews that
 exhibits both content and sentiment information, it is difficult to
 define an appropriate metric.

 To the best of our knowledge, there are no systems
 that combined sentiment analysis with summarization techniques
 to generate product review summaries. Therefore, we have constructed a system
 that incorporates the results from both sentiment analysis
 and summarization, aiming to fuse the advantages of both tasks.
 
\section{Proposed Method} \label{sec:Proposed Method}

\subsection{Motivation} \label{subsec:Preliminaries}
 In order to justify the need to discover the
 underlying reasons in users' opinions, we first compare
 between the outputs of existing product review summarization systems
 such as Google Product\footnote{http://www.google.com/products/, as of 2010},
 Bing Shopping\footnote{http://www.bing.com/shopping/, also as of 2010},
 Hu's system \cite{Hu-KDD04}, and the output that
 we aim to produce in our system.  

\begin{figure}[t]
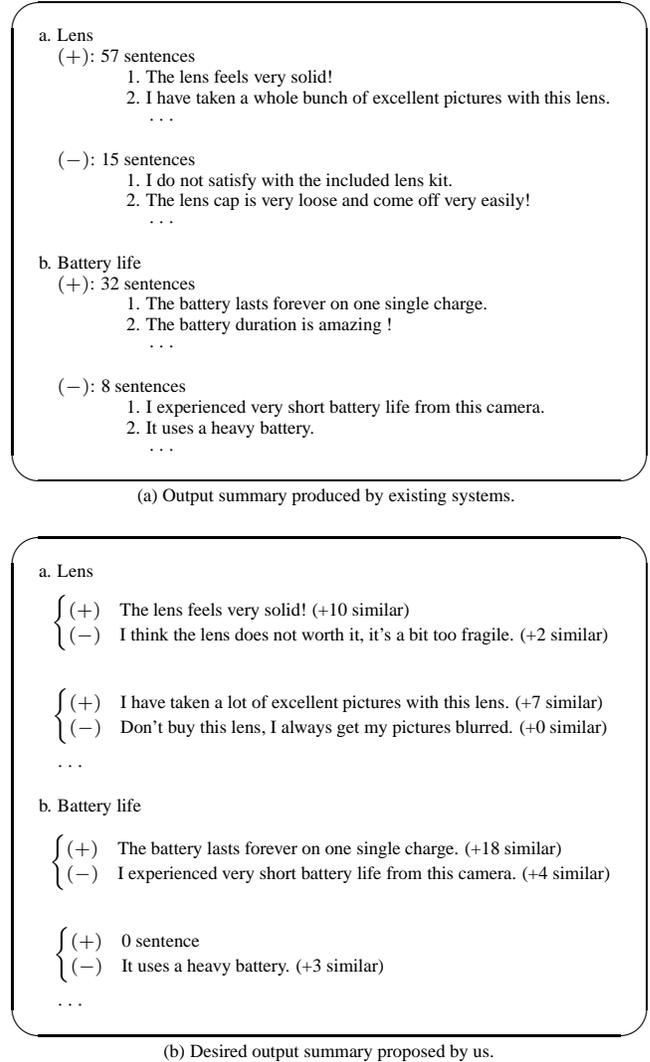

\begin{screen}
\scriptsize
a. Lens \\
$~~~~$ $(+)$: 57 sentences \\
$~~~~~~~~~~~~~~~~~~~~$1. The lens feels very solid! \\
$~~~~~~~~~~~~~~~~~~~~$2. I have taken a whole bunch of excellent pictures with this lens. \\
$~~~~~~~~~~~~~~~~~~~~~~~~\cdots$ \\ \\
$~~~~$ $(-)$: 15 sentences \\
$~~~~~~~~~~~~~~~~~~~~$1. I do not satisfy with the included
 lens kit. \\
$~~~~~~~~~~~~~~~~~~~~$2. The lens cap is very loose and
 come off very easily! \\
$~~~~~~~~~~~~~~~~~~~~~~~~\cdots$ \\

\noindent
b. Battery life \\
$~~~~$ $(+)$: 32 sentences \\
$~~~~~~~~~~~~~~~~~~~~$1. The battery lasts forever on one single
 charge. \\
$~~~~~~~~~~~~~~~~~~~~$2. The battery duration is amazing ! \\
$~~~~~~~~~~~~~~~~~~~~~~~~\cdots$ \\ \\
$~~~~$ $(-)$: 8 sentences \\
$~~~~~~~~~~~~~~~~~~~~$1. I experienced very short battery life from
 this camera. \\
$~~~~~~~~~~~~~~~~~~~~$2. It uses a heavy battery. \\
$~~~~~~~~~~~~~~~~~~~~~~~~\cdots$
\end{screen}
\centering
\scriptsize
(a) Output summary produced by existing systems.
\vspace{3mm}

\begin{screen}
a. Lens
\begin{equation}
 \begin{cases}
  (+) & \text{The lens feels very solid! (+10 similar)} \\
  (-) & \text{I think the lens does not worth it, it's a bit too fragile. (+2 similar)}
 \end{cases}
\nonumber
\end{equation}

\begin{equation}
 \begin{cases}
  (+) & \text{I have taken a lot of excellent pictures with this lens. (+7 similar)} \\
  (-) & \text{Don't buy this lens, I always get my pictures blurred. (+0 similar)}
 \end{cases}
\nonumber
\end{equation}

$\quad\cdots$ \\
 
b. Battery life
\begin{equation}
 \begin{cases}
  (+) & \text{The battery lasts forever on one single charge. (+18 similar)} \\
  (-) & \text{I experienced very short battery life from this camera. (+4 similar)}
 \end{cases}
\nonumber
\end{equation}

\begin{equation}
 \begin{cases}
  (+) & \text{0 sentence} \\
  (-) & \text{It uses a heavy battery. (+3 similar)}\quad\quad\quad\quad\quad\quad\quad\quad\quad\quad\quad\quad
 \end{cases}
\nonumber
\end{equation}
$\quad\cdots$
\end{screen}
(b) Desired output summary proposed by us.
\caption{Comparison of summmaries obtained from
 (a) existing, and (b) our proposed systems.} 
\label{fig:1}
\end{figure}

 Figure~\ref{fig:1} shows two summaries, one that represents
 the existing systems, as well as one that represents our target output.
 Both summaries are structured 
 naturally based on product facets.
 However, the summary
 in Figure~\ref{fig:1}(a) provides only the total number of
 positive and negative sentences (($+$) and ($-$), respectively)
 for each facet, and there is no attempt to organize
 the sentences shown below the number.
 We see that users still need to review  the (possibly numerous) individual
 sentences to discover the actual set of reasons that
 justify the given sentiment. Therefore, it does not satisfy
 the ultimate purpose of a summary. To address this, as illustrated in
 Figure~\ref{fig:1}(b), a summary that provides reasoning of the likes and dislikes
 is preferable, as it maks such direct information explicit.

 The reader may question that the proposed summary is similar to
 Figure~\ref{fig:1}(a) in structure. but simply with an additional level of subtopics.
 Here, we do point out that Figure~\ref{fig:1}(b) is not just a finer grained 
 version of Figure~\ref{fig:1}(a). The grouping of subtopics provides
 a good form of {\it reasoning} and {\it indication} to users on
 what facets ({\it e.g.}, lens, battery life) are liked/disliked.

\subsection{System Overview} 
\label{subsec:SystemOverview}
\begin{figure*}[t]
\centering
 \includegraphics[scale=0.41]{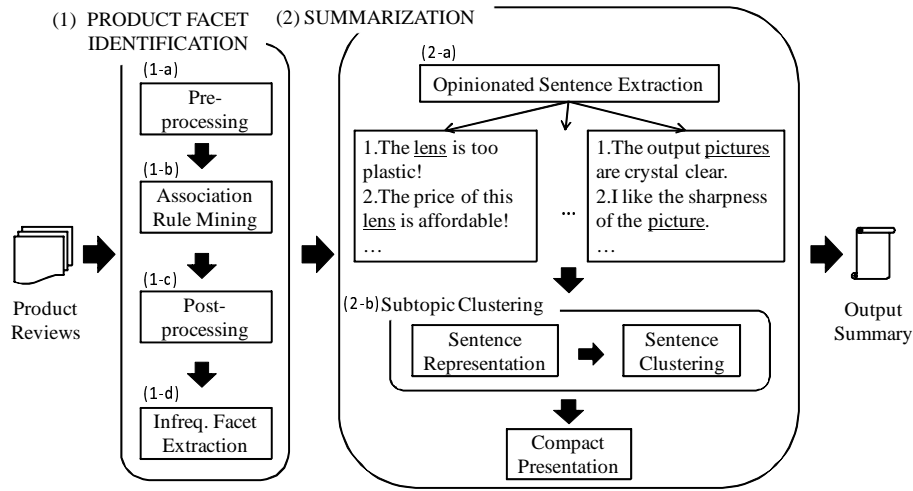}
\caption{System overview.}
\label{fig:System}
\end{figure*}

 Figure~\ref{fig:System} shows an overview of our product review
 summarization system. Our system consists of two main components: \\
 (1) product facet identification, and (2) summarization.

 Aside from the text of the review itself, a review may also feature 
 additional information such as date, time, title
 author name and star-based ratings. For inputs to the
 (1) product facet identification component,
 we do not use any of these information sources, relying on only
 the text body alone, so that our approach is most general.
 We first preprocess these sentences with a Part-of-Speech (POS)
 tagger to obtain the POS label for each word. In the next step,
 only those words that received the label as
 $Noun$ or $Adjective$ -- being part of noun phrases -- are collected
 and fed to the association
 mining module, which generates a list of candidate
 frequent product facets. This is followed by
 some post processing operations in order to remove
 redundant results.
 Last but not least,
 all the adjectives associated with those frequent facets
 in the sentences are also gathered, and used as a means to
 look up those infrequent facets.  Finally, opinionated sentences
 that contain product facet are extracted.

 In the (2) summarization component, 
 the input are groups of sentences that belong to
 each of the product facets obtained from the 
 (1) product facet identification component. We preprocess
 this list of facets to identify and remove insignificant facets.
 In the next step, we start considering
 sentences under each facet independently from others. 
 Each group of sentences is sent to the subtopic clustering
 module. This clustering module 
 first defines a ``sentence representation'' based on
 the similarity between any two sentences,
 and then combines similar sentences to generate clusters. 
 The output from this module is fed to
 the ``compact presentation'' module, which applies
 sentiment analysis and summarization techniques
 to generate the final summary.

\subsection{Product Facet Identification} \label{subsec:PFI}

\subsubsection{Assumptions} 

 It is important to justify that we follow the same assumption
 described in \cite{Hu-KDD04}, so that we consider only
 product facets that appear as nouns or noun phrases; our method
 has the limitation that it cannot handle implicit facets that are
 not explicitly mentioned.
 To explain this crucial point,
 suppose the following two sentences from camera reviews:	
 \begin{itemize}
   \item[(1)] The pictures of this camera are very clear.
   \item[(2)] The camera fits nicely into my palm.
 \end{itemize}
 In the sentence (1), the user expresses his/her satisfaction about the
 quality of the picture taken by the camera, and we can infer that
 the noun \textit{picture} is a facet of the camera. On the other hand,
 the sentence (2) discusses the size of the camera. However,
 the word \textit{size} does not appear explicitly in the sentence.
 In order to identify implicit product facets,
 we need deep semantic understanding of the domain, which implies
 that we have to rely on algorithms that have semantic knowledge of words, a difficult level of technology at the present time. Fortunately,
 explicit facets appear more often in the reviews than
 implicit ones. In our implementation, we consider 
 a span of continuous words as a noun phrase when its rightmost word is a noun
 and the rest of the phrase is composed of nouns or adjectives
 (\textit{e.g.}, battery life, external flash). 
 
\subsubsection{Preprocessing}
\label{subsubsec:Heuristics}

\noindent
\textbf{Part-of-Speech Tagging}

 We utilize the Stanford POS
 Tagger\footnote{http://nlp.stanford.edu/software/tagger.shtml}~\cite{Toutanova00}
 to process each input sentence and yield the part-of-speech (POS)
 label for each word. We observe that the tagger performs fairly well
 at identifying the correct label for nouns and noun phrases, even though
 there are a number of oddly-structured sentences present
 in the reviews. We do not consider stopwords in the tagging
 results, while the remaining noun and noun phrases are also
 converted to their stemmed version using the Porter
 stemmer\footnote{http://www.tartarus.org/\~{}martin/PorterStemmer/}
 \cite{Porter80}.  
The following shows a sentence
 ``I recommend this camera for excellent picture quality'' 
 with the POS tag ($NN$ and $JJ$ are labels for noun and adjective
 respectively): \\

\noindent
I$/PRP$ recommend$/VB$ this$/DT$ camera$/NN$ for$/IN$ \\
excellent$/JJ$ picture$/NN$ quality$/NN$ .\\

\noindent
\textbf{Syntactic Roles}

 We need to further refine the performance of our module  in terms of
 precision by filtering away noisy results. For instance,
 the following words are all accepted as candidate product facets
 when we process a set of camera reviews: ``light,'' ``hand,''
 ``time,'' ``month,'' ``hour,'' and so on. While these nouns often
 appear in the reviews, they are not pruned by any of
 the statistical criteria employed in Hu and Liu's
 system \cite{Hu-KDD04}. Therefore,
 we introduce the use of syntactic roles within a sentence as a feature
 to help distinguish a genuine product facet from such noisy
 ones. Consider the following sentences
 parsed by Stanford Dependency
 Parser\footnote{http://nlp.stanford.edu/software/lex-parser.shtml}~\cite{Klein03}:
 \begin{itemize}
  \item[(1)] The larger \textit{lens} of the \textit{g3} gives better \textit{picture quality} in low light.
  \item[\hspace{5 mm}] \ldots, nsubj(gives-7, lens-3), \ldots, dobj(gives-7, quality-10), \ldots
  \item[(2)] When I took outdoor \textit{photos} with plenty of light, the \textit{photos} were awesome.
  \item[\hspace{5 mm}] \ldots, dobj(took-3, photos-5), \ldots, nsubj(awesome-14, photos-12), \ldots
  \item[(3)] My fiance just did not like the \textit{size}, it is so small in her hand.
  \item[\hspace{5 mm}] \ldots, dobj(like-6, size-8), \ldots
 \end{itemize}
 According to the examples above, we observe that
 genuine facets tend to appear as either subjects or objects
 within the sentences. In fact, our analysis on a subset of
 camera reviews (more than 300 sentences that contain some
 facets over 24 reviews) shows that more than 90\%
 of the instances correspond to the above observation.

 This is not too surprising 
 as subjects and objects in the sentences are usually the targets
 at which the users express their opinions.
 These findings suggest that we can filter non-subject and non-object
 nouns and noun phrases  from the set of identified candidate facets.
 Compared with the processing pipeline
 in Hu and Liu's system~\cite{Hu-KDD04}, we introduce our own heuristic
 during the preprocessing step so that only those legitimate
 noun or noun phrases are delivered to the association mining step, in addition to the
 infrequent facet extraction step where the system does not extract
 those noun or noun phrases that do not appear above a certain number of times.

\subsubsection{Association Rule Mining}
\label{subsubsec:ASRM}
 In this component, we use association rule mining
 technique~\cite{Agrawal94} to statistically identify
 all the frequent explicit product facets. Before we draw the relation
 between association rule mining and our domain of interest,
 we outline the general descriptions of this technique as follows: \\

\vspace{-1mm}
\noindent
\textbf{Items:} \\
 An item is the smallest entity being considered in a particular domain of
 interest. An itemset is a set of items, and the set of all items is
 denoted as $I$. \\

\vspace{-1mm}
\noindent
\textbf{Transaction:} \\
 Transaction $t$ contains itemset $X$ if $X\subseteq t$. The set of
 all transactions is denoted as $D$.\\

\vspace{-1mm}
\noindent
\textbf{Association Rule:} \\
 $X\Rightarrow Y$ where $X\subseteq I$, $Y\subseteq I$ and $X\cap Y=\emptyset$\\

\vspace{-1mm}
\noindent
\textbf{Support:} \\
 $supp(X)$ is the number of transactions in $D$ that contain itemset
$X$. If applied to a rule, $supp(X\Rightarrow Y)=supp(X\cup Y)$. \\

\vspace{-1mm}
\noindent
\textbf{Confidence:} \\
 $cond(X\Rightarrow Y)$ is the number of transactions in $D$ that contain
 itemset $X$ if only contain itemset $Y$. \\

\noindent
 The mining of association rules is then defined as generating
 all possible rules that have support and confidence greater than
 the user-defined minimum values. The Apriori algorithm~\cite{Agrawal94}
 solves this using the following two phases: (i) Identify all frequent
 itemsets that satisfy the minimum support, and (ii) Generate rules
 from those discovered frequent itemsets that satisfy the minimum confidence.

 When we apply this algorithm to our approach, the items are the nouns and noun
 phrases extracted from the ``Preprocessing'' step and the transactions are the
 sentences containing those nouns and noun phrases. We only need to run
 the first phase of the Apriori solution in order to obtain the set of
 frequent itemsets, or equivalently the set of candidate
 frequent product facets. At the same time, we also conveniently obtain
 the ranking for this set of candidate frequent product facets 
 based on their support values. This ranking is an important aspect
 that we utilize in the downstream summarization module when presenting
 information to the users.

\subsubsection{Post Processing}
\label{subsubsec:post-prcs}
 As we consider a large portion of possible nouns and noun phrases
 appeared in the review, not all are genuine facets;
 \textit{i.e.}, some of them are not interesting or redundant. Therefore,
 post processing step removes those irrelevant
 facets by applying the following rules: \\

\noindent
\textbf{Usefulness Pruning}

 This criterion focuses on removing single-word facets that are
 likely to be meaningless. For example, in the context of
 camera reviews, \textit{life} itself is not a useful
 facet, while \textit{battery life} is a meaningful facet.
 We can solve this problem by computing the pure support of
 a facet $f$, which is defined as the number of sentences
 that $f$ appears alone without being
 subsumed by any other facets. If this number is below a predefined
 threshold, there is a strong evidence that we can just keep
 the superset of $f$  as the useful facet. \\

\noindent
\textbf{Compactness Pruning}

 This criterion targets redundant facet phrases --
 noun phrases that are discovered as facets. For example,
 \textit{photo pixel}, \textit{sample image} are not
 as compact as \textit{pixel} and \textit{image}.
 For each of words that the phrase contains, we compute
 the ratio between the support of the phrase and the support of
 that individual word. If any of these ratios is less than
 predefined threshold, we prune the facet phrase. 

\subsubsection{Infrequent Facet Extraction}
\label{subsubsec:infreq-facet-ext}
 As stated thus far, association mining is not able to discover
 infrequent product facets, as they have fairly low support
 value. However, in the case of product facets, users tend to put
 similar opinion words. To illustrate this fact, let us examine the
 following two sentences:
\begin{itemize}
 \item[(1)] The camera takes absolutely amazing pictures.
 \item[(2)] The accompanied software is amazing.
\end{itemize}
 In Sentence (1), \textit{picture} is a frequent facet that has been
 identified by our association mining module, while \textit{software}
 in Sentence (2) is an infrequent one, and thus rejected by frequency.
 On the other hand, we observe that they have the common adjective
 \textit{amazing}.  Hence, our heuristic works in the following two
 steps: (i) gather all opinion words that modify frequent facets; (ii)
 if a sentence contains an infrequent facet candidate, but is modified
 by one or more of the opinion words from (i), the nearest noun and
 noun phrase is included as a facet.
 In this way, we can recover ``software'' as a product facet.

\subsection{Summarization} \label{subsec:Summarization}

\subsubsection{Opinionated Sentence Extraction} \label{subsubsec:Sim-Pruning}
 Sentences that contain any of the product facets that
 we have discovered are labeled with that corresponding facet.
 A sentence can be assigned to more than one facet, as that sentence may
 discuss a relation between many facets. The following instances
 show sentences being labeled with one and two product facets respectively:
\begin{itemize}
 \item[(1)] The \textbf{lens} blocks the \textbf{viewfinder} when the lens
	    is set to wide angle.
 \item[(2)] The 10 \textbf{megapixels} produces really sharp \textbf{pictures}.
\end{itemize}
 It is important to note that we do not feed all labeled sentences into the summarization
 component. We choose opinionated sentences only, since we place larger
 emphasis on summarizing users' opinions in this work.
 In order to achive this, we apply the technique of sentiment analysis
 to filter the labeled sentences based on the approach proposed
 in~\cite{Ding08}: we first prepare a seed list of known-polarity
 adjectives using synonym/antonym pointers in WordNet, and cover
 the other unknown adjectives. The sentence polarity is
 then determined as the summation of all subjectivity scores of
 those adjectives in the sentence. If the resulting summation score is
 positive (negative), the sentence is classified
 as positive (negative). \\

\noindent
\textbf{Similarity Pruning} 

 Users can also employ synonyms to mention the same facet.
 For example: \textit{picture} versus \textit{image}, \textit{photo};
 or \textit{screen}
 versus \textit{monitor}. However, they are treated as different
 genuine facets in Hu and Liu's system~\cite{Hu-KDD04}.
 If we follow this definition,
 different pieces of summary for the same facet will be produced,
 which is not desirable. To solve this problem, we apply
 Kong \textit{et al.}'s word semantic similarity measure~\cite{Kong07}
 to compute the similarity between any of two candidate facets.
 If the score is greater than a predefined threshold, the two words
 (and hence their correspondent sentences) are combined together.

 Kong \textit{et al.}~\cite{Kong07} constructed an edge-counting based model
 that considers the depth of least common subsumer
 and the shortest path length between any two words
 in WordNet. Formally, given two words
 $w_{1}$ and $w_{2}$, the semantic similarity $s_{w}(w_{1},w_{2})$
 is defined by Equation~(\ref{eq:1}):
\begin{equation}
s_{w}(w_{1},w_{2}) = \frac{f(d)}{f(d) + f(l)}, \label{eq:1} 
\end{equation}
 where $l$ is the length of the shortest path
 between $w_{1}$ and $w_{2}$,
 $d$ is the depth of the least common subsumer
 in the WordNet hierarchical semantic net,
 and $f(x)$ denotes the transfer function for $d$ and $l$.
 For $s_{w}(w_{1},w_{2})$, the interval of similarity is $[0,1]$,
 $1$ for the maximum similarity and $0$ for no similarity at all.
 We follow the experimental results shown in~\cite{Kong07}
 and choose $f(x) = e^{x} - 1$. The resulting formula is:
\begin{equation}
 s_{w}(w_{1},w_{2}) = \frac{e^{\alpha d}-1}{e^{\alpha d} + e^{\beta l} - 2} \;\;\; (0 < \alpha, \beta < 1), \label{eq:2}
\end{equation}
 where $\alpha$ and $\beta$ are smoothing factors.
 As reported in~\cite{Kong07}, the optimal values of
 $\alpha$ and $\beta$ are both 0.25. 
 We also use these optimal values in our experiments. \\

\noindent
\textbf{Sentence Representation and Similarity Measurement} 

 After identifying product facets, sentences 
 are analyzed to determine
 their subjectivity. 
 To facilitate the subsequent clustering algorithm,
 we decide to adopt a simple yet novel sentence representation,
 together with a sentence similarity measurement scheme
 proposed in~\cite{Li06}, which yields state-of-the-art results.
 At a high-level view, the algorithm utilizes a dynamic
 vector representation that adapts to the size of the sentence,
 and computes the cosine similarity between two sentence vectors.

 The algorithm starts with identifying ``concepts''
 in the sentence \cite{Ye05}. Concepts are defined as those open class
 words (nouns, verbs, adjectives and adverbs, excluding stopwords)
 in the sentence. We additionally employ the restriction on syntactic roles,
 described in Section~\ref{subsubsec:Heuristics} so that we only include
 those words that hold subject and object roles
 in the sentence. In detail, we extract important nouns
 that are subject or object, main verbs/adjectives associated with
 those important nouns, adverbs that modify the main verb/adjectives.
 Then given two sentences for which we want to compute similarity, $s_{1}$
 with the set of concepts $C_{1}$, and $s_{2}$ with the set of
 concepts $C_{2}$, we define a  joint concept vector 
 $C = C_{1} \cup C_{2}$. In the next step,
 $V_{i}$ -- the vector representation for $s_{i}$ ($i=1,2)$ --
 is created, with size equal to that of $C$,
 whose values are determined by the following rules: \\

\noindent
 At index $k$, 
\begin{itemize}
 \item	If $s_{i}$ contains $C[k]$ -- concept at $k^{th}$ index
 in the joint vector, $V_{1}[k]$ is set to $1.0$.
 \item	If $s_{i}$ does not contain $C[k]$, a semantic similarity score
	is computed between $C[k]$ with all concepts in that
	sentence. $V_{i}[k]$ is then set to the highest similarity
	score.
 We apply the same Equation~(\ref{eq:2}) to compute similarity.
\end{itemize}

 The semantic similarity between two sentences $s_{1}$ and $s_{2}$
 can now be measured by the cosine similarity between the two
 representative vectors 
 $\boldsymbol{V_{1}}$ and $\boldsymbol{V_{2}}$, respectively,
 which results in a score within the range $[0,1]$.
 This similarity is defined by Equation~(\ref{eq:3}):
\begin{eqnarray}
 sim(s_{1}, s_{2})=\frac{\boldsymbol{V_{1}} \cdot \boldsymbol{V_{2}}} {\left\| \boldsymbol{V_{1}} \right\|\cdot \left\| \boldsymbol{V_{2}} \right\| }. \label{eq:3}
\end{eqnarray}	
 Figure \ref{fig:sentence representation} shows an example of the above steps
 for clarification.
\begin{figure}[t]
\begin{screen}
\footnotesize
Assume that the following two sentences with the underlined concepts:
\begin{eqnarray}
s_{1}&=&\text{The \underline{battery} of this camera is very \underline{impressive}.} \nonumber \\
s_{2}&=&\text{Canon \underline{camera} always \underline{has} a \underline{long} \underline{battery} \underline{life}.} \nonumber
\end{eqnarray}
 Therefore, the joint vector is denoted as follows:
\begin{eqnarray}
 C=\text{\{battery, camera, impressive, has, long, life}\} \nonumber
\end{eqnarray}
 The resulting sentence vectors $V_{1}$ and $V_{2}$ are as follows:
\begin{eqnarray}
 V_{1}&=&\{1.0, 1.0, 1.0, 0.0, 0.3, 0.15\} \nonumber \\
 V_{2}&=&\{1.0, 1.0, 0.3, 1.0, 1.0, 1.0\} \nonumber
\end{eqnarray}
 The semantic similarity between two sentences, $s_{1}$ and $s_{2}$
 is computed as follows:
\begin{eqnarray}
 sim(s_{1}, s_{2})=0.69 \nonumber
\end{eqnarray}
\end{screen}
\caption{Example of sentences together with their vector representation.}
\label{fig:sentence representation}
\end{figure}

\subsubsection{Subtopic Clustering for Summarization}
 Once all pairwise similarities are  calculated,
 we feed the set to the sentence clustering module. We implemented
 both hierarchical and non-hierarchical algorithms to compare
 their performances. \\

\noindent
\textbf{(1) Hierarchical Clustering}

 We apply hierarchical clustering in an agglomerative (bottom-up)
 manner. Individual sentences are initialized as singleton clusters,
 and are iteratively merged to form clusters with the minimum pairwise
 distance together.  This continues until a terminating criterion is
 satisfied.  The well-known pairwise cluster distances are
 complete-link, single-link and groupwise-average.  Among them, we
 employ groupwise-average distance as our preliminary experimentation
 shows that it performs more consistently. Given two different
 clusters $c_{i}$ and $c_{j}$, the groupwise-average distance is
 defined as follows:
\begin{eqnarray}
\label{eq:4} 
 &sim(c_{i}, c_{j})& \nonumber \\
&=\frac{1}{\left|c_{i}\cup c_{j}\right| \; \left(\left|c_{i}\cup c_{j}\right| - 1\right)}&\sum_{x \in c_{i} \cup c_{j}} \sum_{y \in c_{i} \cup c_{j} : y \neq x} sim(x,y). \nonumber
\end{eqnarray}
 Too many small clusters result in an excessively detailed summary
 and an over-estimation of the number of actual subtopics, while a few large
 clusters result in a summary that omits important information.
 Therefore, we adopt an algorithm proposed in \cite{Hatz01}
 to estimate the final number of clusters. The clustering process will
 terminate as soon as the number of clusters exceeds this value.
 In \cite{Hatz01}, they first defined the notion of links:
 if the semantic similarity score between any two sentences are
 greater than a certain threshold, a link is posited, joining the two 
 sentences together.
 Therefore, if we compute the similarity score for every two sentences
 in the collection and apply the notion of links, a graph
 with the vertex being sentences, and edges representing those links
 will be created. Then the number of estimated clusters $c$
 given the input of $n$ sentences that correspond to a graph with $m$
 connected components is defined as follows:
\begin{equation}
 c = m + \left(\frac{n}{2} - m\right)\left(1 - \frac{\log(L)}{\log(P)}\right),
\label{eq:5}
\end{equation}
 where $L$ is the observed number of links. In addition,
 the maximum possible number of links $P$ is defined as follows:
\begin{eqnarray}
 P=\frac{n(n-1)}{2}. \nonumber
\end{eqnarray}

\noindent
\textbf{(2) Non-hierarchical Clustering}

 We also implement a non-hierarchical clustering technique,
 the exchange method~\cite{Spath85}, which regards
 the clustering problem as an optimizing task. The algorithm
 seeks to minimize an objective function $\Phi$ that measures
 the intra-cluster dissimilarity between a partition
 $P=\{C_{1}, C_{2}, \cdots , C_{k}\}$:
\begin{eqnarray}
 \Phi(P)=\sum_{i=1}^{k}\left(\frac{1}{|C_{i}|}\sum_{x,y\in C_{i},x\neq y}
(1-sim(x,y))\right). \label{eq:nh-cls}
\end{eqnarray}
 The same estimation on the number of final clusters mentioned
 earlier is first applied to determine the size of the partition $P$.
 The algorithm then proceeds by creating an initial assignment
 of the sentences into the partition, and looking for locally
 optimal moves (``swaps'') of sentences between clusters 
 that improve $\Phi$ in each iteration until convergence.
 Since this is a hill-climbing method, it is necessary to call
 the algorithm multiple times, with random partition of sentences
 into the clusters each time. The optimal overall configuration
 will be selected as the final clustering result. \\

\noindent
\textbf{(3) Compact Presentation of Sentences}

 This step generates and presents the resulting target summary
 shown in Figure~\ref{fig:1} (b). It considers sentence
 clusters from all facets generated by the previous
 ``Subtopic Clustering'' component. 
 By applying the sentiment analysis technique described in
 Section~\ref{subsubsec:Sim-Pruning}, we can determine the orientation
 for every sentence in a particular subtopic. With this information,
 we are able to partition the sentences in each subtopic based on
 their polarity. The subsequent task is to select the most
 representative sentence for each partition.  The selected sentence
 must represent the maximum information present in the other
 sentences; in other words, the target sentence is most similar to all
 the remaining sentences. Thus, we define a metric to compute the
 representative power of a sentence as follows:

 For each sentence $s_{i}$ in the correspondent positive/negative
 partition $P$, we define its representative power $Rep(s_{i})$ 
 as follows:
\begin{eqnarray}
 Rep(s_{i})=\sum_{s_{j}\in P-s_{i}} sim(s_{i},s_{j}).
\end{eqnarray}
 The sentence with the highest representative power will be selected
 as the output sentence to users.  Finally, for the user's quick reference,
 we also supplement the selected sentence with 
 the number of sentences sharing the same point of view.

\section{Experiments} \label{sec:Experiments}

\subsection{Experimental Data and Measure} \label{subsec:ExpDataMeasure}

\subsubsection{Experimental Data} 

 In our experiments, we use publicly available sets of reviews for
 three products (camera, phone, and DVD) \cite{Hu-KDD04}. This dataset is
 directly compatible to our ``product facet identification''
 component, since we evaluate our implemented version of
 Hu and Liu's system and our proposed system
 in the exactly the same way as in \cite{Hu-KDD04}. In addition,
 to evaluate the summarization component, we prepare
 our own labeled data, which consists of sentences
 being partitioned into subtopics for a set of
 22 most frequent facets extracted from those three products.
 The inter-annotator agreement between two annotators was 85\%.
 The final extraction of the data for evaluation
 that reached both annotators' consensus was 90\%. 

\subsubsection{Evaluation Measure for Product Facet Identification} 

 We use the standard precision and recall measures to evaluate
 the performance of our product facet identification
 component. Let $MF$ and $SF$ be manually extracted facets
 and system extracted facets, respectively. Precision $(Pre)$
 and recall $(Rec)$ are defined as follows:
\begin{eqnarray}
 \label{eq:7} 
 Pre = \frac{\left| \left\{MF\right\} \cap \left\{SF\right\} \right|}{\left| \left\{SF\right\}\right|}, \quad
 Rec = \frac{\left| \left\{MF\right\} \cap \left\{SF\right\} \right|}{\left| \left\{MF\right\}\right|}. \nonumber
\end{eqnarray}

\subsubsection{Evaluation Measure for Summarization} 
 In order to evaluate the performance of our summarization component,
 we use purity, inverse purity, and $F$-measure (the harmonic mean of
 purity and inverse purity) that are widely used
 clustering measures \cite{Hotho05}. 

 Purity is related to the precision measure. This measure
 focuses on the frequency of the most common category in each cluster,
 and rewards the clustering algorithm that introduce less noise
 in each cluster. Let $C$, $L$, and $n$ be
 the set of automatic clusters to be evaluated,
 the set of manual annotated clusters,
 and the number of sentences to be clustered, respectively.
 Purity is computed by taking the weighted average of maximum precision values:
\begin{equation}
	\label{eq:9} 
 Purity = \sum_{i}\frac{\left|C_{i}\right|}{n}\max \text{Precision}(C_{i},L_{j}),
\nonumber
\end{equation}
 where the precision of an automatic cluster $C_{i}$
 for a given manual subtopic $L_{j}$ is defined as:
\begin{equation}
\label{eq:10} 
\text{Precision}(C_{i}, L_{j}) = \frac{\left|C_{i} \cap L_{j}\right|}{\left|C_{i}\right|}. \nonumber
\end{equation}
\begin{table*}[t]
\centering
\caption{Performance of the product facet identification component in
 Hu and Liu~\cite{Hu-KDD04}.}
\label{Table:1}
\begin{tabular}{c|c|c|c|c|c|c|c}\hline\hline
Data & Number of manually & \multicolumn{2}{c|}{Association mining} &
 \multicolumn{2}{c|}{Post processing} & \multicolumn{2}{c}{Infrequent facet}
 \\ \cline{3-8} 
     & extracted facets & Recall & Precision & Recall & Precision &
 Recall & Precision \\ \hline
Camera  & 79            & 0.671  &  0.552    & 0.658  & 0.825	  &
 0.822  & 0.747 \\ \hline
Phone & 67 & 0.731 & 0.563 & 0.716 & 0.828 & 0.761 & 0.718 \\ \hline
DVD & 49 & 0.754 & 0.531 & 0.754 & 0.765 & 0.797 & 0.793 \\ \hline
Average & 65 & 0.719 & 0.549 & 0.709 & 0.806 & 0.793 & 0.753 \\ \hline\hline
\end{tabular}
\end{table*}

\begin{table*}[t]
\centering
\caption{Performance of our product facet identification component, comprising of 
 Hu and Liu's system \cite{Hu-KDD04} + the use of syntactic roles.}
\label{Table:2}
\begin{tabular}{c|c|c|c|c|c|c|c}\hline\hline
Data & Number of manually & \multicolumn{2}{c|}{Association mining} &
 \multicolumn{2}{c|}{Post processing} & \multicolumn{2}{c}{Infrequent facet}
 \\ \cline{3-8} 
     & extracted facets & Recall & Precision & Recall & Precision &
 Recall & Precision \\ \hline
Camera & 79 & 0.671	& 0.646 & 0.658	& 0.894	& 0.822	& 0.842 \\ \hline
Phone & 67 & 0.731 & 0.648 & 0.716 & 0.903 & 0.761 & 0.769 \\ \hline
DVD & 49 & 0.754 & 0.610 & 0.754 & 0.818 & 0.797 & 0.867 \\ \hline
Average & 65 & 0.719 & \textbf{0.634} & 0.709 & \textbf{0.872} & 0.793 & \textbf{0.826} \\ \hline\hline
\end{tabular}
\end{table*}

 Inverse Purity focuses on the cluster with maximum recall
 for each category, rewarding clustering solutions
 that gather more elements of each category
 in a corresponding single cluster. Inverse Purity ($I$-$Purity$) is
 defined as follows:
\begin{equation}
\label{eq:11} 
I\text{-}Purity = \sum_{i}\frac{\left|L_{i}\right|}{n}\max \text{Precision}(L_{i},C_{j}). \nonumber
\end{equation}

 The $F$-measure $F_{\alpha}$ that is the harmonic mean of
 purity and inverse purity is also defined as follows:
\begin{equation}
\label{eq:12} 
F_{\alpha} = \frac{1}{\alpha \frac{1}{\text{Purity}} +  (1-\alpha) \frac{1}{\text{Inverse Purity}}}. \nonumber
\end{equation}
 In our evaluation, we set the value of $\alpha$
 to 0.5, and denote it as $F_{1}$ (rather than $F_{0.5}$ to follow
 standard $F_1$ semantics) in the following.

\subsection{Experimental Results} \label{subsec:ExpResults}

\subsubsection{Product Facet Identification} 
 Tables~\ref{Table:1} and ~\ref{Table:2} show
 the results of our implemented version of
 Hu and Liu's system~\cite{Hu-KDD04},
 and the results when we integrate
 heuristic of syntactic roles into their system,
 respectively.
 Table~\ref{Table:1} shows that our reimplementation
 can achieve the results reported in~\cite{Hu-KDD04}.
 We observe that the system identifies most of
 the common facets such as \textit{battery},
 \textit{picture}, \textit{lens} for camera,
 \textit{signal}, \textit{headset} for phone
 and \textit{remote control}, \textit{format}
 for DVD player.
 We observe an 
 improvement in precision in Table~\ref{Table:2}
 as most of noisy results have been filtered away
 using syntactic role information.
 For example, in \textit{Camera} dataset,
 while the precision in infrequent facet extraction
 in Table~\ref{Table:1} achieves 0.747,
 the precision, infrequent facet extraction
 in Table~\ref{Table:2} achieves 0.842. This shows
 0.095 improvement.
 However, we observe no improvement in recall
 since the syntactic role heuristic is a filter, eliminating noise
 rather than adding new results.

\subsubsection{Summarization} 
 Table~\ref{fig:Table:3} shows the results for
 the summarization component. Each of facets contains
 different number of subtopics, even as low as one.

 For example, the $Price$ facet in the $DVD$ product actually has no
 subtopic, resulting in just one manually defined cluster. The reason
 is that users only express their opinions toward two extremes on
 whether the DVD player is expensive or affordable
 (note that subtopic is independent of sentiment information).
 Similarly, for the $Format$ facet in the $DVD$ product,
 users only discuss whether the DVD player can play
 all video formats or not. Thus, the number of manually defined
 clusters is also one. 

 On the other hand, some facets
 have a lot of subtopics ({\it e.g.}, $Lens$ in Camera (7 subtopics),
 $LCD$ in Camera (6 subtopics), {\it etc.}).  This is due to the fact that
 they exhibit many different properties
 (the size, ease of use, price, {\it etc.} for the $lens$,
 or the resolution, material, color, {\it etc.} for $LCD$).
 Users do discuss the many angles of
 these subtopics. We also observe that the common facet
 $Service$ in $Phone$ produces more subtopics (5 subtopics)
 compared with those mentioned in $DVD$ (1 subtopic). This is because
 generally, $Phone$ users tend to compare among many different
 service providers, while $DVD$ users only complain
 about the service of that particular manufacturer in the review,
 with almost no comparison to its competitors.
\begin{table*}[t]
\centering
\small
\caption{Performance of the Summarization component.}
\label{fig:Table:3}
\begin{tabular}{c|c|c|c|c|c|c|c|c|c|c|c}\hline\hline
Data & Facet & Number of manually & \multicolumn{3}{c|}{\textbf{Hierarchical
 clustering}} & \multicolumn{3}{c|}{\textbf{Non-hierarchical clustering}} & \multicolumn{3}{c}{\textbf{Random clustering}} \\ \cline{4-12}    
     &       & defined clusters & $Purity$ & $I$-$Purity$ & $F_{1}$ & $Purity$
 & $I$-$Purity$ & $F_{1}$ & $Purity$ & $I$-$Purity$ & $F_{1}$  \\ \hline
     & Battery & 4 & 0.864	& 0.591	& 0.702 & 0.864 & 0.636 & \textbf{0.733} & 0.864	& 0.455 & 0.596 \\ \cline{2-12}
     & Memory & 3 & 0.643	& 1.000	& \textbf{0.783} & 0.643 & 0.786 & 0.707 & 0.500	& 0.643 & 0.563 \\ \cline{2-12}
     & Flash & 4 & 0.556	& 0.722	& 0.628 & 0.667 & 0.722 & \textbf{0.693} & 0.500	& 0.611 & 0.550 \\ \cline{2-12}
     & LCD & 6 & 0.478	& 0.826	& 0.606 & 0.565 & 1.000 & \textbf{0.722} & 0.348 & 0.739 & 0.473 \\ \cline{2-12}
Camera & Lens & 7 & 0.792	& 1.000	& \textbf{0.884} & 0.792 & 1.000 & \textbf{0.884} & 0.500	& 0.667 & 0.571 \\ \cline{2-12}
     & Megapixels & 5 & 0.621	& 0.483	& 0.543 & 0.724 & 0.552 & \textbf{0.626} & 0.552	& 0.414 & 0.473 \\ \cline{2-12}
     & Mode & 6 & 0.813	& 1.000	& \textbf{0.897} & 0.813 & 1.000 & \textbf{0.897} & 0.500	& 0.625 & 0.556 \\ \cline{2-12}
     & Shutter & 6 & 0.643	& 0.929	& \textbf{0.760} & 0.643 & 0.929 & \textbf{0.760} & 0.429	& 0.786 & 0.555 \\ \cline{2-12}
     & Average & 5.13 & 0.676	& 0.819	& 0.725 & 0.714 & 0.828 & \textbf{0.753} & 0.524	& 0.617 & 0.542 \\ \hline\hline
     & Battery & 3 & 0.824	& 0.765	& \textbf{0.793} & 0.765 & 0.706 & 0.734 & 0.706	& 0.588 & 0.642 \\ \cline{2-12}
     & Camera & 3 & 0.727	& 0.636	& \textbf{0.679} & 0.727 & 0.636 & \textbf{0.679} & 0.727	& 0.545 & 0.623 \\ \cline{2-12}
     & Headset & 4 & 0.467	& 0.733	& \textbf{0.570} & 0.400 & 0.600 & 0.480 & 0.400	& 0.667 & 0.500 \\ \cline{2-12}
     & Radio & 3 & 0.737	& 0.737	& \textbf{0.737} & 0.737 & 0.737 & \textbf{0.737} & 0.737	& 0.579 & 0.648 \\ \cline{2-12}
Phone & Service & 5 & 0.438	& 0.875	& 0.583 & 0.563 & 1.000 & \textbf{0.720} & 0.375	& 0.625 & 0.469 \\ \cline{2-12}
     & Signal & 3 & 0.824	& 0.941	& \textbf{0.878} & 0.824 & 0.765 & 0.793 & 0.824	& 0.588 & 0.686 \\ \cline{2-12}
     & Size & 3 & 0.760 & 0.680	& 0.718 & 0.920 & 0.680 & \textbf{0.782} & 0.720	& 0.520 & 0.604 \\ \cline{2-12}
     & Speaker & 4 & 0.684	& 0.895	& \textbf{0.775} & 0.684 & 0.789 & 0.733 & 0.684	& 0.632 & 0.657 \\ \cline{2-12}
     & Average & 3.50 & 0.682	& 0.783	& 0.717 & 0.702 & 0.739 & \textbf{0.722} & 0.647	& 0.593 & 0.604 \\ \hline\hline
     & Price & 1 & 1.000 & 0.714	& 0.833 & 1.000 & 0.762 & \textbf{0.865} & 1.000	& 0.524 & 0.688 \\ \cline{2-12}
     & Remote & 4 & 0.625	& 0.750	& \textbf{0.682} & 0.563 & 0.750 & 0.643 & 0.500	& 0.688 & 0.579 \\ \cline{2-12}
     & Format & 1 & 1.000	& 0.714	& \textbf{0.833} & 1.000 & 0.571 & 0.727 & 1.000	& 0.500 & 0.667 \\ \cline{2-12}
DVD  & Design & 1 & 1.000	& 1.000	& \textbf{1.000} & 1.000 & 1.000 & \textbf{1.000} & 1.000	& 1.000 & 1.000 \\ \cline{2-12}
     & Service & 1 & 1.000	& 0.739	& \textbf{0.850} & 1.000 & 0.522 & 0.686 & 1.000	& 0.522 & 0.686 \\ \cline{2-12}
     & Picture & 4 & 0.800	& 0.850	& \textbf{0.824} & 0.800 & 0.850 & \textbf{0.824} & 0.450	& 0.500 & 0.474 \\ \cline{2-12}
     & Average & 2.00 & 0.904	& 0.795	& \textbf{0.837} & 0.894 & 0.743 & 0.791 & 0.825	& 0.622 & 0.682 \\ \hline\hline
\end{tabular}
\end{table*}

 Interestingly, the number of subtopics varies not only 
 from facet to facet, but also from product to product.
 In our data, the product $Camera$ shows the greatest
 number, about 5 subtopics per facet on average,
 while $DVD$ only contains 2 subtopics per facet on average.
 This can be explained from the above observation:
 the facets that belong to $Camera$ usually have richer
 properties to be commented on compared with those belong to
 $DVD$. Interestingly, this also impacts
 the performance of our clustering algorithm.

 We compare the performance of our algorithms with a baseline,
 which randomly assigns sentences to clusters. Note that
 the number of clusters is determined by the estimation
 in Equation~(\ref{eq:5}), before the clustering process starts.   The estimated cluster number is fed to the random algorithm as well (for comparison). We record
 the average performance of the random clustering baseline
 over 200 trials.
 For the non-hierarchical clustering
 approach, we also execute the algorithm 200 times,
 in order to ameliorate the effect of occurrences where the algorithm is trapped in a local minimum. 
 We record the run that minimizes the objective function
 in Equation~(\ref{eq:nh-cls}) the best.  However, we need to
 execute the hierarchical clustering algorithm only once,
 as it is a deterministic algorithm given the estimated number of
 final clusters.

 The last row in each product data in Table~\ref{fig:Table:3}
 shows the relative performance of the proposed algorithms
 with respect to the baseline of random clustering.
 According to Table~\ref{fig:Table:3}, 
 our two proposed clustering algorithms always outperform
 the baseline of random clustering by a significant amount.

 On the other hand, we observe small differences in the average
 performance between the hierarchical approach and
 the non-hierarchical one. The non-hierarchical approach
 tends to perform better when the number of subtopics is large
 ({\it e.g.}, $Lens$ in Camera, $Service$ in Phone), but performs 
 worse when the number of subtopics is small
 ({\it e.g.}, $Service$ in DVD). An analysis shows that
 when  more subtopics exist, the non-hierarchical
 approach has a better chance to reach the global solution
 as every move/swap operation it suggests affects
 the objective function. However, when we have small number of
 subtopics, its move/swap operation is not as effective,
 and the algorithm also terminates quickly;
 while the hierarchical approach using average-link distance
 keeps a better balance between the clusters. 

 We have shown that both hierarchical and non-hierarchical 
 clustering outperform the baseline of random clustering
 in all three products, Camera, Phone, and DVD.
 However, we observe that the marginal percentage in performance
 between them tends to decrease as the number of
 subtopics reduces. In most cases,
 with a reliable sentence similarity measurement,
 the estimated number of final clusters is indeed very close to
 the annotated subtopics. When we have only a few topics,
 the estimated number of final clusters is also small.
 Under this condition, each sentence assigned by the random
 clustering algorithm also has a higher chance of
 assigning the correct cluster. As a result, 
 we do not observe a large improvement for our proposed clustering
 algorithms over the random algorithm. On the other hand,
 if we have many topics, the estimated number of final clusters
 also becomes larger. This is why the random assignment gets
 little success in assigning sentences to the correct clusters.

\section{Conclusion} \label{sec:Conclusion}
 In this work, we have proposed a system that can summarize
 product reviews.  Existing systems related to
 product reviews summarization usually constructed
 a facet-based summary, which can aggregate
 sentiment information that belongs to each facet.
 We have implemented this similar method as the first component
 in our system. We improve this component's performance by applying
 syntactic role information within a sentence.

 More importantly, since we showed the existence of
 underlying subtopics within facets,
 we introduced a second task that actually summarizes
 the reviews from a deeper perspective. Our summarization component
 proceeded by grouping sentences about the same subtopics together,
 and provided a compact summary with the sentiment information
 to the users. We introduced a clustering approach
 to solve the subtopic problem.
 Nevertheless, the approach is highly dependent on
 the semantic similarity between words as well as sentences,
 which is a problem that we cannot completely solve
 without some forms of manual input. In addition,
 we do not utilize deep semantic information in determining
 the similarities between sentences. If we are able to analyze
 such semantics, our system may be able to  achieve better performance.

 Several extensions from our current system are possible.
 Different brand names that belong to a particular product
 class (\textit{e.g.}, Nikon, Canon (Camera); Pioneer (DVD);
 iPod (Music Player), \textit{etc}.), or product/manufacture names
 of the accessories that go together with the main product
 (\textit{e.g.}, Kingston (compact flash card for camera),
 Nvidia (graphic card for computer, \textit{etc}.),
 are all treated as genuine facets
 in the annotation from the dataset. However,
 in most cases, they appear together with some other
 facets when comparison is made between that product
 and its competitors (``My Canon camera has longer
 battery life than Nikon''). These general/proper entities are
 not very useful for summarization and should be excluded.
 It is one of the future works to build a module
 that recognizes these proper names and excludes them.
 Comparative-based summarization system would
 benefit directly from our systems, as it is now able to
 compare product facets at a more fine-grained level. Alternatively,
 as our summarization system only generates extractive-based summary,
 it might be more desirable to have a system that can reformulate
 the output sentences from our subtopic clustering
 and provides users with content. Last but not least,
 more useful metadata about the reviews such as title,
 users' ratings, and so on can also be augmented to
 the summarization system.

\bibliographystyle{abbrv}
\bibliography{./KhangJCDL11refs} 

\end{document}